\documentclass[9pt, conference]{IEEEtran}
\usepackage{algpseudocode}    % new algorithm package
\usepackage{algorithm}
\usepackage{amsmath}
\usepackage{amssymb}
\usepackage{amsfonts}
\usepackage{booktabs}
\usepackage{multirow}
\usepackage{subfigure}
\usepackage{graphicx}         % include pdf figures
\usepackage{color}
\usepackage{cite}             % more citations in one bracket
\usepackage{comment}          % use comment
\usepackage{soul}             % use highlight command \hl{}
\soulregister\cite7
\soulregister\ref7
\soulregister\pageref7
\usepackage{amsthm}
\usepackage{etoolbox}         % commands \newtoggle, \toggletrue, \iftoggle
\usepackage{url}
\usepackage{nth}              % nth command
\usepackage{bm}               % bm command

\makeatletter
\let\OldStatex\Statex
\renewcommand{\Statex}[1][3]{%
  \setlength\@tempdima{\algorithmicindent}%
  \OldStatex\hskip\dimexpr#1\@tempdima\relax
}
\makeatother

\graphicspath{{./figs/}}
 
\usepackage{ulem}
\usepackage{color}
\usepackage{xcolor}
\usepackage{pifont}
\usepackage{makecell}
\usepackage{markdown}
\usepackage{tcolorbox}
\usepackage{amsmath}
\usepackage{graphicx}

\usepackage{xurl} % 允许 URL 在任意地方换行
\usepackage[hidelinks]{hyperref} % 超链接
\usepackage{svg}
\svgpath{{figs/}}

\usepackage{booktabs}
\usepackage{array}
\usepackage{makecell}

\title{
\textbf{
Simulation-Aware In-Context Policy Improvement for LLM-Aided Analog Layout Refinement
}
}
\author{
    \IEEEauthorblockN{Bingyang Liu, Ziming Wei, Xiaohan Gao, and David Z. Pan}
    \IEEEauthorblockA{Department of Electrical \& Computer Engineering, The University of Texas at Austin, TX, USA \\
    \{bingyangliu@, zmwei@, xiaohan.gao@austin., dpan@ece.\}utexas.edu}
}

\begin{document}
\IEEEoverridecommandlockouts
\IEEEpubid{\makebox[\columnwidth]{979-8-3195-1246-8-0/26/\$31.00 \copyright2026 IEEE \hfill}
\hspace{\columnsep}\makebox[\columnwidth]{ }}
% 2 column?

% \input{doc/cover}
% \input{doc/answer}

% % remove redaundant content for ACM template
% \settopmatter{printacmref=false}
% \setcopyright{none}
% \renewcommand\footnotetextcopyrightpermission[1]{}

% 注释掉这条去掉页码
% \pagestyle{plain}

\maketitle

\begin{abstract}
    % ICLAD compacted
    Analog IC layout design remains a labor-intensive iterative process dominated by simulation-driven refinement.
    Although end-to-end layout generators accelerate initial placement and routing, they still require experts to manually tune layout optimization parameters with repeated post-layout simulations for stringent design specifications.
    While Bayesian Optimization (BO) is widely adopted for parameter tuning in analog IC design, at the layout level it typically requires hundreds to thousands of evaluations, each involving costly parasitic extraction and post-layout simulation, which makes it impractical.
    Recently, Large Language Models (LLMs) have demonstrated potential in improving the sample efficiency of such simulation-driven tuning. However, their restricted access to geometric layout context and design-specific heuristics limits their ability to manipulate the layout optimization process. In this paper, we propose a simulation-aware LLM multi-agent framework that performs in-context policy improvement (ICPI) by iteratively updating layout optimization parameters exposed by an analog layout generator through an act–observe–reflect loop on compact structured layout representations. Experiments on real-world analog circuits show that, with only tens of post-layout simulations, our approach improves post-layout performance over the generator's built-in heuristics and BO-based tuning method.
\end{abstract}

% \begin{IEEEkeywords}
%   Large language model, Multi-agent, Analog layout design, Interactive layout editing.
% \end{IEEEkeywords}

\section{Introduction}
\label{sec:Introduction}

% placement is important 
% what is placement 
% state-of-the-art placement 
% previous parallelization effort 
% our contribution 

\begin{sloppypar}
Analog IC layout design is a highly iterative, simulation-driven process. To reach design-specific performance targets, designers repeatedly modify layouts and run costly post-layout simulations, and industrial flows still rely on experienced designers to iteratively refine layouts from sparse simulation feedback. Because this refinement is a matter of reasoning over feedback rather than following a fixed procedure, it has stayed a human-driven loop, carried out by expert designers.
\end{sloppypar}

\begin{sloppypar}
In academia, decades of research on analog placement and routing (P\&R) algorithms and automated layout generators have accelerated initial layout generation, from classical formulations~\cite{391116,510537,malavasi1990routing,xiao2010practical,xu2017hierarchical} and domain-knowledge-driven methods~\cite{basaran1993latchup,ou2013simultaneous,martins2016current,xu2019device,ho2013coupling} to machine-learning techniques~\cite{li2020exploring,li2020customized,gusmao2020semi}. Recent end-to-end analog layout generators including ALIGN~\cite{kunal_align_2019,sapatnekar_align_2023}, MAGICAL~\cite{xu_magical_2019,chenMAGICALOpenSource2021,chen_magical_2021}, and other frameworks~\cite{gao2024joint,zhang2023sage,zhang2024sage} optimize wirelength-centric analytical objectives augmented with heuristics and manually specified constraints. However, the real post-layout performance is far more complex than what these analytical surrogates capture. Since traditional automated flows typically apply generalized, static heuristics to all designs, they often fail to capture the flexible trade-offs required for specific high-performance targets. Therefore, designers still need to manually tune layout optimization parameters through repeated simulations, which limits the practical adoption of such tools.
% However, the real post-layout performance model is much more complex than these analytical surrogates. Designers still need to iteratively run simulations and manually tune the constraints and hyperparameters to obtain design-specific, high-quality layouts, which limits automation and adoption. 
\end{sloppypar}

\begin{figure}[tb]
    \centering
    \includegraphics[width=0.48\textwidth]{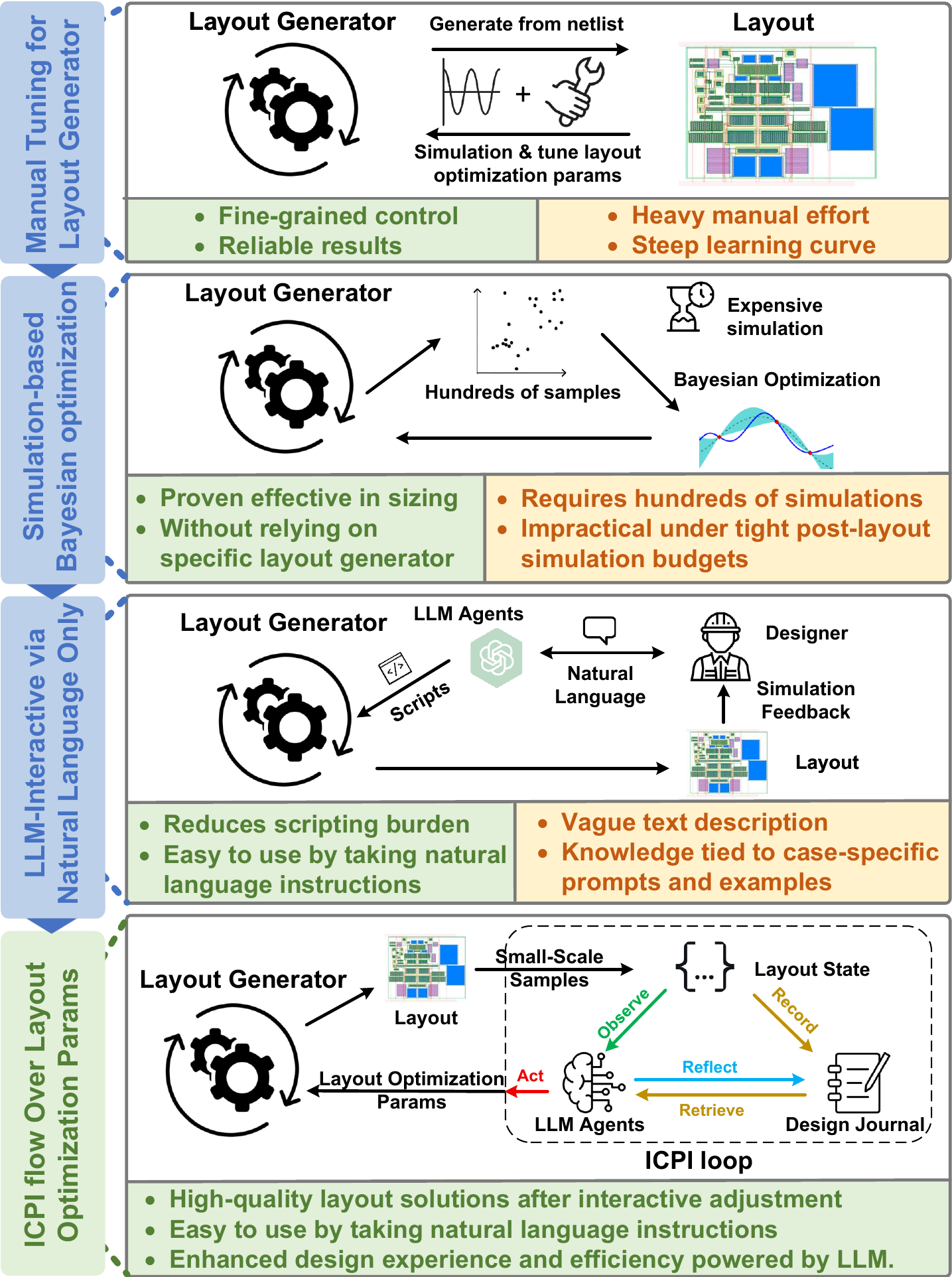}
    \caption{Comparison of four simulation-driven analog layout flows built on end-to-end layout generators: manual tuning, simulation-based Bayesian optimization, LLM-interactive design via natural language only, and our ICPI flow over layout optimization parameters.}
    \label{fig:Intro}
    %\vspace{-0.7cm}
\end{figure}

\begin{sloppypar}
A natural question is whether algorithms can automatically explore better settings of layout optimization parameters in analog layout generators based on simulation feedback. Simulation-based tuning with Bayesian Optimization (BO) has been widely used for analog sizing tasks~\cite{yin2022fast,budak2021efficient,liu2021parasitic,du2022surrogate,turbo,kong2024pvtsizing}. However, BO typically requires hundreds to thousands of samples per design, which is impractical for layout, where each sample involves costly parasitic extraction and post-layout simulation. The simulation budget is often limited to only tens of runs per design, which falls far short of the sampling requirements for effective exploration through BO.
\end{sloppypar}

\begin{sloppypar}
Recent work has explored leveraging the reasoning capabilities of Large Language Models (LLMs) to improve the sample efficiency of such simulation-based methods~\cite{kochar2025ledro,yin2024ado,liu2025llm,ahmadzadeh2025anaflow, wei2025toposizing}, suggesting that LLMs might help make better use of scarce evaluations. Compared with front-end sizing, back-end layout offers a particularly suitable setting for LLMs: automated layout generators including MAGICAL~\cite{xu_magical_2019,chenMAGICALOpenSource2021,chen_magical_2021} expose semantically meaningful layout optimization parameters within their placement and routing algorithms, such as symmetry constraints and routing priorities. These structured parameters naturally align with the ability of LLMs to reason about circuit structure and layout trade-offs. However, two challenges still hinder their use for layout design automation. First, the LLM must perceive geometric layout changes through a faithful, evolving state representation rather than vague textual descriptions. Second, it must learn to adjust layout optimization parameters under a tight post-layout simulation budget. Prior LLM-powered efforts for analog layout design~\cite{liu2024layoutcopilot,wang2024chatpattern} largely rely on natural-language descriptions of layout as input, which can be inherently ambiguous. They also inject domain knowledge mainly through prompt engineering and case-specific examples that often need to be rebuilt in new designs.
A separate line of recent work orchestrates the full intent-to-layout flow with LLM agents, where constraint-driven generation produces the layout and post-layout metrics are fed back to select the best implementation among complete attempts~\cite{zhang2026panda}. Because such flows maintain neither an explicit layout state nor accumulated design-specific experience that guides how feedback should update layout optimization parameters, the two challenges above remain open there as well.
These unresolved challenges make it difficult to substantially and reliably improve analog layout generation in a simulation-aware and scalable manner. As summarized in Fig.~\ref{fig:Intro}, current simulation-driven flows therefore rely on manual tuning, simulation-based BO, or natural-language-only LLM interaction on top of such generators, each with its own limitations under tight post-layout simulation budgets.
\end{sloppypar}

\begin{sloppypar}
In this paper, we introduce a simulation-aware multi-agent framework that performs in-context policy improvement (ICPI) by iteratively updating a set of exposed layout optimization parameters. Here, layout optimization parameters refer to generator-exposed placement and routing parameters, including net weights, placement bias, symmetry constraints, routing priorities, and wire widths. The framework treats layout refinement as an adaptive process that evolves a design-specific policy for each circuit instance under a fixed budget of tens of post-layout simulations. It exposes current geometry, constraints, and metrics through a compact, LLM-friendly intermediate representation, the layout state, and retains simulation-verified experience in a design journal for retrieval and reflection across rounds, all without updating model weights. In each iteration, the agents select one family of layout optimization parameters to adjust, propose concrete edits, observe feasibility and post-layout feedbacks, and record the resulting reflections as new experience. Our main contributions are summarized as follows:
\end{sloppypar}

\begin{sloppypar}
    \begin{itemize}
        \item We design a compact, LLM-friendly layout state that contains structured, multi-level information including circuit connectivity, device locations, exposed layout optimization parameters, parasitic summaries, and post-layout simulation outcomes. This structured state replaces purely natural-language layout descriptions and enables state-aware reasoning during refinement.
        \item We build a simulation-aware act--observe--reflect ICPI loop around a mature analog layout generator, exposing a structured set of layout optimization parameters for placement and routing refinement. In each round, the agents update one parameter family based on the current layout state and sparse post-layout feedback, while storing validated experience in a persistent design journal.
        \item We validate the framework on two OTA benchmarks across different technologies and scales of the parameter space. Under the small budget of post-layout simulations, our method achieves better post-layout performance than the generator's built-in heuristics, BO-based tuning, and a non-ICPI baseline.
    \end{itemize}

    % DAC
    % \begin{itemize} 
    %     \item We design a compact, LLM-friendly layout state that exposes circuit connections, device locations, current net-level control policies, net-level parasitics, and post-layout simulation results. This structured representation replaces purely natural-language descriptions, reduces ambiguity in layout description. 
    %     \item We build an act–observe–reflect ICPI loop with LLM agents around a mature analog P\&R kernel, exposing a set of net-level controls as the action space. In each round, the agents update one family of controls based on current layout state and simulation results, and writes validated experiences to a persistent design journal. 
    %     \item We validate our framework on two OTAs in different technology, demonstrating better post-layout performance than automated generators with default net-level control, Bayesian optimization based method, and non-ICPI prompting under equal simulation budgets.
    % \end{itemize}
\end{sloppypar}

\begin{sloppypar}
The rest of the paper is organized as follows. 
Section~\ref{sec:Preliminary} describes the background,
Section~\ref{sec:Algorithm} explains the detailed algorithm and implementation, 
Section~\ref{sec:Results} demonstrates the results, and 
Section~\ref{sec:Conclusion} concludes the paper. 
\end{sloppypar}

\section{Preliminaries}
\label{sec:Preliminary}

% This section reviews the background concepts of our study, including the integration of LLMs with EDA, prompt engineering, multi-agent collaboration, and the interactive placement and routing in analog layout design,  additionally outlining the scope of \textsl{LayoutCopilot}.

\subsection{In-Context Policy Improvement and Self-Feedback}

\begin{sloppypar}
Recent work on the self-feedback and internal consistency of LLMs shows that they can improve their behavior at test time by structuring feedback, reflection, and retrieval around their own outputs. Self-refinement methods~\cite{madaan2023self, chen2023teaching} follow a propose–feedback–refine pattern: the model first produces an answer, then generates natural-language feedback or confidence signals, and finally rewrites or abstains based on this feedback, yielding quality gains under fixed weights. Beyond purely in-context refinement, self-improvement methods further exploit self-generated signals for training, including rationales, scores, and language feedback~\cite{huang2023large, patel2024large}. Liang \textit{et al}.~\cite{liang2024internal} further unify these approaches under a common view where behavior is improved by exploiting self-generated evaluation signals. Here, we use the term in-context policy improvement (ICPI) to refer specifically to the test-time variant with frozen weights, where the ``policy'' is realized by prompts, retrieved memories, and structured context rather than by parameter updates.
\end{sloppypar}

\begin{sloppypar}
ICPI ideas also appear in multi-step agents and domain-specific optimization. ReAct-style agents~\cite{react} interleave chain-of-thought reasoning with environment actions, and Reflexion~\cite{shinn2023reflexion} augments this loop with explicit self-feedback, where the model summarizes failures and proposes strategy updates that are fed into later prompts. Similar propose–execute–revise patterns underlie code optimization and configuration frameworks~\cite{jiang2023selfevolve,yu2025autonomous,isr_llm}, in which the model repeatedly edits programs or settings, runs tests or benchmarks, and uses execution traces to guide further changes. Across these settings, performance improves because the in-context state becomes richer over iterations, including past trajectories, feedback, and heuristic rules.
\end{sloppypar}

% \begin{sloppypar}
% ICPI ideas have also been extended to multi-step agents and domain-specific optimization. ~\cite{reflexion} execute episodes in an environment, obtain scalar or structured task feedback, and write verbal reflections that analyze failures and propose strategy adjustments; these reflections are stored and injected into prompts for future episodes. Similar patterns appear in code optimization frameworks~\cite{selfevolve,seidr,satlution,alphaevolve} and other domain specific frameworks~\cite{isr_llm, repower,improve}, where models iteratively modify programs or configurations, run test suites or benchmarks, and use execution traces to guide further edits. In all these cases, improvement arises because the context including past trajectories, reflections, and rules becomes richer across iterations, which incorporates experience enabling LLM to behave better in the environment.
% \end{sloppypar}

\begin{sloppypar}
These studies suggest that ICPI is particularly well suited to tasks where the model’s actions correspond to semantically meaningful controls and the environment returns rich, domain-specific signals each round. Analog layout back-end optimization exhibits both properties: automated generators expose interpretable layout optimization parameters including symmetry constraints, wirelength weights, routing priorities, and wire widths and spacings, while parasitic extraction and post-layout simulation provide high-value though costly feedback. This makes simulation-aware ICPI well suited to our setting and motivates the structured layout state, design journal, and iterative scheme introduced in this work.
\end{sloppypar}

\subsection{End-to-end Analog Layout Generator}

\begin{sloppypar}
Modern end-to-end analog layout generators build on decades of research on analog placement and routing algorithms to produce layouts directly from circuit netlists, technology information, and user constraints. Systems such as MAGICAL~\cite{xu_magical_2019,chenMAGICALOpenSource2021,chen_magical_2021}, ALIGN~\cite{kunal_align_2019,sapatnekar_align_2023}, and recent analog P\&R frameworks~\cite{gao2024joint,zhang2023sage,zhang2024sage} share a broadly similar pipeline: given a schematic and high-level constraints, they first place devices, then legalize the placement to satisfy design rules while preserving key relations, and finally run an analog-aware router that enforces symmetry, spacing, and other heuristics. This integrated place–legalize–route flow systematically encodes many analog layout requirements into analytical objectives, substantially reducing manual effort for producing a DRC-clean first-pass layout.
\end{sloppypar}

\begin{sloppypar}
Despite this progress, most practical flows still rely on designers to provide and tune constraints and tool-level parameters, including symmetry groups, optimizer configurations, wire spacing options, and a small set of global weights that balance different objectives. In principle, analog layout generators could expose much finer-grained net-level optimization parameters, including initial device positions or anchors, well-cluster assignments, per-net wirelength weights, routing width and spacing options, and per-net routing priorities. However, the large number of such layout optimization parameters makes systematic manual tuning impractical. As a result, these settings often remain fixed across simulation iterations and even across designs, which leaves a large portion of the design space underexplored. In this work, we modify a place-and-route kernel derived from the open-source analog layout generator MAGICAL~\cite{xu_magical_2019,chenMAGICALOpenSource2021,chen_magical_2021} to explicitly surface a structured subset of such net-level layout optimization parameters, serving as the action space for the ICPI framework described next.
\end{sloppypar}

\begin{figure*}[tb]
    \centering
    \includegraphics[width=0.96\textwidth]{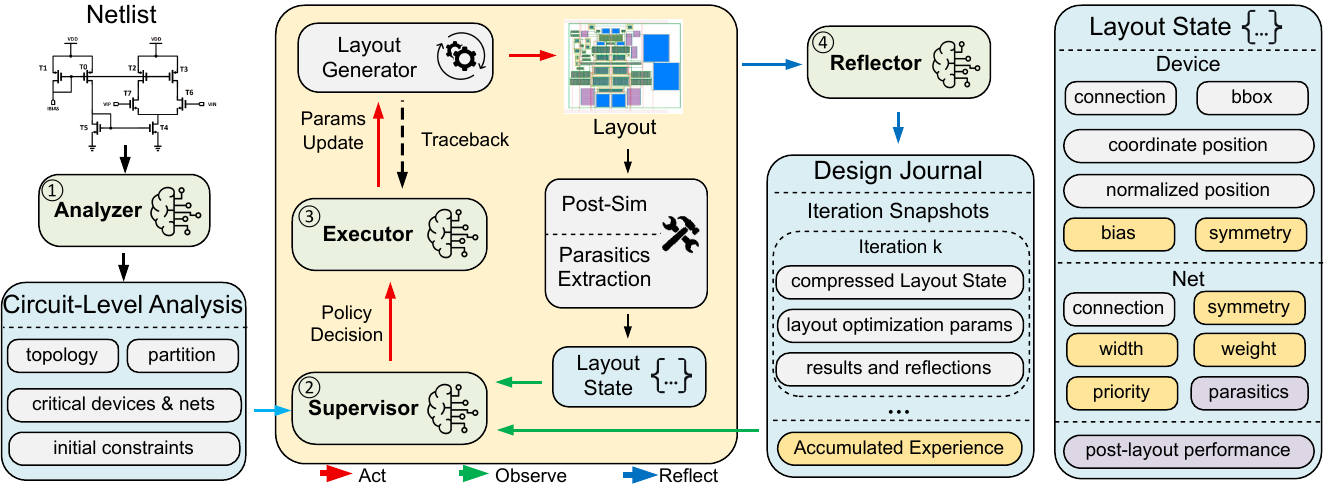}
    % \caption{Overview of the proposed LLM multi-agent ICPI framework. Light-green rounded boxes are LLM agents, and the yellow region highlights the act--observe--reflect loop: green arrows indicate observing layout and simulation feedback, red arrows indicate acting on exposed controls through the P\&R kernel, and blue arrows reflect experience back into the design journal. Three panels denote structured information, purple labels mark post-layout simulation feedback, and yellow labels mark exposed P\&R kernel control families updated during the loop, as summarized in Table~\ref{tab:actionspace}.}
    \caption{Overview of the proposed LLM multi-agent ICPI framework. Light-green boxes represent LLM agents executing an act--observe--reflect loop within the yellow region. Colored arrows denote observation (green), action (red), and reflection (blue) steps. The right panels illustrate the structured layout state and design journal, where color-coded labels distinguish simulation feedback (purple) from the parameter families (yellow) detailed in Table~\ref{tab:actionspace}.}
    \vspace{-0.4cm}
    \label{Fig2.Framework}
\end{figure*}

\section{Framework}
\label{sec:Algorithm}

\begin{sloppypar}
As discussed in Section~\ref{sec:Introduction}, we focus on two challenges in LLM-aided simulation-aware analog layout refinement: (i) how to give the LLM precise access to the evolving layout beyond ambiguous natural-language descriptions, and (ii) how to improve decisions over layout optimization parameters across iterations under a budget of only tens of expensive post-layout simulations. As shown in Figure~\ref{Fig2.Framework}, we address these challenges through three core components: a structured \emph{layout state} that synchronizes the current layout and feedback, a persistent \emph{design journal} that stores design-specific experience across rounds, and a simulation-aware act--observe--reflect loop driven by three agents, namely a Supervisor, an Executor, and a Reflector. In addition, a one-shot Analyzer provides circuit-level prior hints to prune the search space before iterative refinement. Table~\ref{tab:actionspace} summarizes the exposed parameter families that define the action space of this ICPI loop.
\end{sloppypar}

\subsection{Action Space and Circuit-Level Analysis}
\label{subsec:enviroment}

% v5
\begin{table}[tb]
\caption{Layout optimization parameter families used in ICPI.}
\label{tab:actionspace}
\footnotesize
\setlength{\tabcolsep}{6pt}
\renewcommand{\arraystretch}{1.15}
\begin{tabular}{@{}>{\raggedright\arraybackslash}m{0.24\linewidth}
                   >{\raggedright\arraybackslash}m{0.66\linewidth}@{}}
\toprule
\textbf{Param.\ Family} & \textbf{Effect} \\
\midrule
Net weights    & Increase the weight of selected nets in the placement objective. \\
\addlinespace
Placement bias & Bias selected devices toward preferred layout directions in placement. \\
\addlinespace
Symmetry       & Enforce symmetry constraint on selected devices during placement and legalization. \\
\addlinespace
Priority       & Prioritize routing on critical nets. \\
\addlinespace
Wire widths    & Trade off parasitic resistance and capacitance during routing. \\
\bottomrule
\end{tabular}
\end{table}

\begin{sloppypar}
Our framework runs on top of an analog layout generator derived from the open-source layout generator MAGICAL~\cite{xu_magical_2019,chenMAGICALOpenSource2021,chen_magical_2021}, with extended PDK support and additional interfaces that expose layout optimization parameters.
We keep the underlying optimization algorithms unchanged, but explicitly expose a structured set of layout optimization parameters that can be updated by the LLM agents during refinement. In this work, we use \textit{layout optimization parameters} to refer to the generator-exposed placement and routing parameters that directly influence layout evolution during optimization, rather than generic tool hyperparameters. As summarized in Table~\ref{tab:actionspace}, these parameters are organized into five interpretable parameter families: net weights, placement bias, symmetry constraints, routing priorities, and wire widths. These parameters have direct geometric or electrical implications, including pulling devices on critical nets closer in placement, biasing selected devices toward preferred directions, tightening matching on sensitive structures, prioritizing routing resources, and trading off parasitic resistance and capacitance through wire width. In the following, we treat this parameter space as the action space of the ICPI loop, while the resulting parameter settings and their geometric consequences are encoded back into the layout state for subsequent rounds.
\end{sloppypar}

\begin{sloppypar}
Although the exposed parameter space is much more structured than raw layout editing, it is still combinatorially large under even coarse discretization, especially when applied across a large number of nets and devices. Under a budget of only tens of expensive post-layout evaluations, purely black-box exploration can cover only a very small fraction of this space. We therefore introduce an off-loop Analyzer that processes the netlist once before iterative refinement. 
% It decomposes the circuit into functional blocks, identifies devices and nets likely to be sensitive to parasitics or mismatch, and provides candidate symmetry structures together with coarse importance hints. These outputs are reused in every round without re-parsing the netlist, allowing the Supervisor and Executor to focus on a pruned subset of controls and likely critical structures. In this sense, the Analyzer serves as a circuit-level prior that improves search efficiency, but the main adaptation during refinement still comes from the iterative ICPI loop itself.
It decomposes the design into stages and functional blocks, identifies critical devices and nets that are sensitive to mismatch or parasitics, and suggests candidate symmetry constraints together with rough importance scores for different devices and nets. Intuitively, this step uses the LLM’s reasoning ability and pretrained circuit knowledge to separate likely critical structures from background details. 
These outputs are reused in every round without re-parsing the netlist, allowing the Supervisor and Executor to focus on a pruned subset of parameters and likely critical structures. In this sense, the Analyzer serves as a circuit-level prior that improves search efficiency, but the main adaptation during refinement still comes from the iterative ICPI loop itself. In practice, this pruning is especially important because the exposed parameters are heterogeneous: some primarily affect placement geometry, while others mainly affect routing parasitics or matching-related constraints. The Analyzer therefore does not attempt to predict the final best setting directly, but instead narrows the search to parameter families and structures that are more likely to matter under sparse post-layout feedback.
\end{sloppypar}

%\vspace{-0.2cm}
\subsection{Layout State}
\label{subsec:state}

\begin{sloppypar}
As discussed earlier, we consider two key challenges. The first is how to convey layout information to the LLM without exposing raw polygons or the full GDS description. Polygon-level representations are verbose and lack explicit circuit structure, while image-based representations fail to capture circuit topology, making it difficult to associate rectangles with devices and to infer routing connectivity. We therefore define a layout state as a structured, synchronized intermediate representation of the current layout, as illustrated in Figure~\ref{Fig2.Framework}. It organizes multi-level information that matters for refinement into two views: a device-level view that captures geometry and placement-related structure, and a net-level view that captures routing-related properties and parasitic summaries, as well as current parameter settings.
\end{sloppypar}

\begin{sloppypar}
Concretely, the layout state combines four types of information. First, it records structural context, including connectivity, subcircuit membership, bounding boxes, and coordinate-based geometric features, so that the LLM can reason about where devices and nets sit geometrically and how they are topologically related. Second, it records the current exposed parameters from Subsection~\ref{subsec:enviroment}, allowing the agents to see which parameter family is currently active and how the layout is being steered. Third, after each placement-and-routing result, it incorporates parasitic extraction summaries that approximately indicate layout quality. Finally, when post-layout simulation is run for the current round, it appends the performance results. By composing structural information, applied parameters, and observed feedback in a single representation, the layout state allows the LLM to answer three questions within a short context window: what the layout looks like now, how it is currently behaving, and which settings produced that behavior.
\end{sloppypar}

\subsection{Design Journal}
\label{subsec:journal}

\begin{sloppypar}
The second challenge is how to let the LLM accumulate design-specific knowledge during optimization, rather than relying only on static prompts or hand-crafted exemplars. Prior LLM-assisted analog layout efforts~\cite{liu2024layoutcopilot,wang2024chatpattern} mainly inject expertise through fixed instructions, and do not maintain persistent memory of which edits helped or hurt within a particular design. In contrast, our goal is to keep the prompts largely design-agnostic and let the model adapt within each design instance through iterative interaction with layout and simulation feedback.
\end{sloppypar}

\begin{sloppypar}
To support this, we introduce a design journal that serves as cross-round memory for the ICPI loop. Each journal entry for round $k$ stores a compressed snapshot of the pre-round layout state, the selected parameter family and intended goal of the round, the concrete edits and outcomes produced by the Executor, and a short reflection on why the update appears beneficial or harmful. It also records practical experience for failure handling, such as repeated routing or legalization breakdowns caused by overly aggressive settings. During subsequent rounds, the journal is retrieved as in-context experience, allowing the agents to reuse previously validated heuristics, avoid repeating unproductive trials, and adapt their parameter decisions to the current design without updating model parameters.
\end{sloppypar}

\subsection{ICPI Loop}

\begin{sloppypar}
The ICPI loop follows an act--observe--reflect pattern driven by three agents, as shown in Figure~\ref{Fig2.Framework}. At the beginning of round $k$, the Supervisor receives the current layout state, the Analyzer’s circuit-level hints, and the design journal. Based on recent parasitic and post-layout feedback, it decides whether the next step should continue refining a previously promising parameter family or explore a different one. It then selects exactly one parameter family $c_k$ from the exposed action space in Subsection~\ref{subsec:enviroment} and formulates a small set of high-level goals for that family, such as strengthening matching around a sensitive pair or prioritizing routing for a set of critical nets. In this way, the Supervisor performs policy-level decision making while keeping attribution at the parameter-family level.
\end{sloppypar}

% \begin{sloppypar}
% The Executor agent focuses on converting the checklist from the Supervisor into concrete updates of the net-level control file. It searches for attempts of the same control family in the design journal to see how changes in that family have historically affected parasitics and post-layout metrics, and uses these examples to choose which specific parameters to update and how far to move. It then proposes a new setting for the selected controls and calls the P\&R kernel. If the kernel fails to produce a legal layout, for instance, due to over-constrained symmetry, overly aggressive width settings, or conflicting well-group assignments, the Executor enters a small inner loop: it analyzes the failure logs, consults the journal entries that record similar failures, and rolls back to safer control values until a feasible solution is obtained. The output of this stage is a new layout and an updated layout state with refreshed parasitics and, when scheduled, post-layout simulation metrics.
% \end{sloppypar}
\begin{sloppypar}
The Executor translates the Supervisor’s goals into concrete parameter updates. It looks up past attempts of the same parameter family in the design journal, uses those experiences to choose which parameters to modify and by how much, and then invokes the layout generator with the proposed settings. If the generator fails to produce a legal layout, for example, because symmetry becomes over-constrained or widths become too aggressive, the Executor enters a small traceback loop that rolls back to safer values using failure logs and related journal entries. The output of this stage is a new legal layout together with an updated layout state containing refreshed parasitic summaries and, when scheduled, post-layout simulation results.
\end{sloppypar}

% \begin{sloppypar}
% Finally, the Reflector updates the design journal with a new entry for round $k$, following the method in Subsection~\ref{subsec:state} and~\ref{subsec:journal}. It summarizes the pre-round layout state, records the Supervisor’s choices and the Executor’s actions and outcomes, and adds a short reflection on why the changes appear beneficial or harmful, which provides experience for subsequent rounds. Over time, the journal evolves into a design-specific instruction, and future rounds can treat it as a library of ``what worked last time in a similar situation’’ when choosing new net-level controls.
% \end{sloppypar}
\begin{sloppypar}
After each round, the Reflector compresses the history into a new journal entry. It summarizes the pre-round state, records the selected parameters and resulting outcomes, and adds a brief reflection on why the change appears helpful or harmful. Over time, this journal becomes a design-specific memory of what has worked, what has failed, and how different parameter families interact with the current circuit. This persistent memory is what turns repeated prompting into in-context policy improvement.
\end{sloppypar}

\begin{sloppypar}
All agents share the same backbone LLM but use role-specific prompts, similar in spirit to Self-Refine~\cite{madaan2023self} and Reflexion-style~\cite{shinn2023reflexion} prompting. The Analyzer receives a structured netlist serialization, whereas the Supervisor receives the current layout state, the Analyzer’s circuit-level hints, and the design journal. Both agents use fixed output schemas, while the Executor is constrained to return only concrete parameter updates and short rationales. The Reflector is prompted to compress the current round into a reusable journal entry. 
\end{sloppypar}

% ---

\section{Experimental Results}
\label{sec:Results}

\subsection{Experimental Setup}

\begin{sloppypar}
We evaluate the proposed framework on two operational transconductance amplifiers (OTAs) from different technology nodes and topologies, which are widely used building blocks in analog design. OTA1 is a two-stage Miller-compensated OTA in a 65\,nm CMOS process, and OTA2 is a fully differential OTA with common-mode feedback in a 40\,nm process. Although the benchmark set is limited to two designs, these cases are not toy examples: each candidate requires a full layout-generation step followed by parasitic extraction, and selected iterations further require post-layout simulation. As a result, the optimization budget is fundamentally different from conventional low-cost parameter search. This makes the benchmark suitable for evaluating whether structured state, circuit-level priors, and design-specific memory can improve search efficiency beyond one-shot prompting and black-box optimization. In both cases, \emph{Gain} denotes the open-loop gain, while UGB and PM are extracted from the closed-loop simulations. For OTA1, the target specifications are \mbox{Gain $\ge 50$\,dB}, \mbox{UGB $\ge 18$\,MHz}, \mbox{CMRR $\ge 80$\,dB}, and \mbox{PM $\ge 60^{\circ}$}. For OTA2, the targets are \mbox{Gain $\ge 60$\,dB}, \mbox{UGB $\ge 1.5$\,MHz}, \mbox{CMRR $\ge 80$\,dB}, and \mbox{PM $\ge 60^{\circ}$}. For both designs, a smaller area is preferred.
\end{sloppypar}

\begin{sloppypar}
Our implementation builds on a layout generator derived from the open-source layout generator MAGICAL~\cite{xu_magical_2019,chenMAGICALOpenSource2021,chen_magical_2021}, with extended PDK support and additional interfaces that expose layout optimization parameters as described in Subsection~\ref{subsec:enviroment}. For each candidate layout, we perform parasitic extraction using Siemens Calibre and post-layout simulation using Spectre within Cadence Virtuoso. The Analyzer, Supervisor, Executor, and Reflector agents all use GPT-5 as the shared backbone LLM, while the framework also supports open-source backbone LLMs through the same interface.
\end{sloppypar}

\begin{sloppypar}
We compare our framework with three baselines. \textit{Heuristic} uses the generator's built-in fixed settings of the layout optimization parameters: all net weights are set to 1, all device bias terms are set to 0, symmetry constraints are manually specified, routing priorities are set to 1 for all nets, and wire widths are set to the minimum width for all non-power nets and to $4\times$ the minimum width for power nets. \textit{BO} applies Bayesian optimization over the same exposed parameter space, treating the post-layout FoM defined below as the objective. \textit{Ours} is the full simulation-aware workflow described in Section~\ref{sec:Algorithm}, with the complete ICPI loop enabled. All optimization-based methods start from the same initial layout-optimization parameters. BO and Ours use the same 31-candidate optimization horizon, with different post-layout simulation schedules as described below. For \textit{Ours}, we run 30 iterations per design. Each iteration invokes the generator and full-circuit parasitic extraction, while post-layout simulation is performed every three iterations, adding up to 31 parasitic extractions and 11 post-layout simulations per design. Because BO requires the objective value at every iteration, we run both extraction and post-layout simulation for all 31 of its iterations. \textit{Ours w/o ICPI} uses the same 31-candidate optimization horizon and post-layout simulation schedule as Ours, but removes the design journal and cross-round reflection. Each update is generated without retrieving experience from previous rounds..
\end{sloppypar}

\begin{sloppypar}
To summarize electrical quality, we define a scalar FoM over $m \in \{\text{Gain},\text{UGB},\text{CMRR}\}$. For each metric with target $L_m$ and value $v_m$, we compute
\begin{equation}
s_m =
\begin{cases}
(\dfrac{v_m}{L_m})^2, & v_m < L_m,\\[4pt]
1 + \dfrac{v_m - L_m}{v_m}, & v_m \ge L_m,
\end{cases}
\end{equation}

\begin{equation}
\alpha_{\mathrm{PM}} = \min\!\left((\dfrac{v_{\mathrm{PM}}}{L_{\mathrm{PM}}})^2,\,1\right)
\end{equation}

% 
% \begin{equation}
% \begin{aligned}
% s_m &=
% \begin{cases}
% \dfrac{v_m}{L_m}, & v_m < L_m,\\[4pt]
% 1 + \dfrac{v_m - L_m}{v_m}, & v_m \ge L_m,
% \end{cases}\\[6pt]
% \alpha_{\mathrm{PM}} &= \min\!\left(\dfrac{v_{\mathrm{PM}}}{L_{\mathrm{PM}}},\,1\right)
% \end{aligned}
% \end{equation}

% embedded
% \[
% s_m =
% \begin{cases}
% \dfrac{v_m}{L_m}, & v_m < L_m,\\[4pt]
% 1 + \dfrac{v_m - L_m}{v_m}, & v_m \ge L_m,
% \end{cases}
% \qquad
% \alpha_{\mathrm{PM}} = \min\!\left(\dfrac{v_{\mathrm{PM}}}{L_{\mathrm{PM}}},\,1\right),
% \]
where $L_{\mathrm{PM}} = 60^{\circ}$. To emphasize violations of minimum performance specifications, we apply a quadratic penalty when a metric falls below its target, while retaining a bounded reward above the target. Thus $s_m = 1$ at the target and increases smoothly when the metric exceeds it, while $\alpha_{\mathrm{PM}}$ caps the contribution of phase margin once the design is sufficiently stable. The final FoM is defined as the geometric mean of $s_{\mathrm{Gain}}$, $s_{\mathrm{UGB}}$, and $s_{\mathrm{CMRR}}$, multiplied by $\alpha_{\mathrm{PM}}$, so FoM is sensitive to degradation in any of the metrics. 
\end{sloppypar}

\subsection{Post-layout Performance Comparison}

\begin{table*}[tb]
\centering
\caption{Post-layout performance comparison on OTA1 and OTA2.}
\label{tab:ota}
\scriptsize
\setlength{\tabcolsep}{3pt}
\renewcommand{\arraystretch}{1.1}
\resizebox{\textwidth}{!}{%
\begin{tabular}{l cccccc cccccc}
\toprule
& \multicolumn{6}{c}{\textbf{OTA1}} & \multicolumn{6}{c}{\textbf{OTA2}} \\
\cmidrule(lr){2-7} \cmidrule(lr){8-13}
Method
& Gain (dB) & UGB (MHz) & CMRR (dB) & PM (deg) & FoM & Area ($\mu\text{m}^2$)
& Gain (dB) & UGB (MHz) & CMRR (dB) & PM (deg) & FoM & Area ($\mu\text{m}^2$) \\
\midrule
Heuristic     & 47.66 & 15.67 & 78.1  & 69.2 & 0.869 & 3592.9
              & 53.15 & 1.454 & 55.70 & 59.4 & 0.696 & 7080.0 \\
BO            & 62.18 & 53.80 & 54.89 & 69.3 & 0.979 & 4036.9
              & 62.14 & 1.610 & 62.80 & 70.4 & 0.880 & 8857.3 \\
Ours w/o ICPI & 50.59 & 17.24 & 87.62 & 69.3 & 1.003 & 3288.3
              & 55.70 & 1.611 & 67.36 & 72.6 & 0.868 & 6490.5 \\
Ours          & 51.39 & 18.69 & 108.6 & 69.4 & \textbf{1.104} & 4031.4
              & 62.72 & 1.597 & 80.86 & 72.3 & \textbf{1.038} & 6468.2 \\
\bottomrule
\end{tabular}
}
\vspace{-0.5cm}
\end{table*}

% \begin{table*}[tb]
% \centering
% \caption{Post-layout performance comparison on OTA1.}
% \label{tab:ota1}
% \scriptsize                      
% \setlength{\tabcolsep}{3pt}      
% \renewcommand{\arraystretch}{1.05}
% \resizebox{0.89\textwidth}{!}{%
% \begin{tabular}{lcccccc}
% \toprule
% Method & Gain (dB) & UGB (MHz) & CMRR (dB) & PM (deg) & FoM & Area ($\mu\text{m}^2$) \\
% \midrule
% Heuristic    & 47.66 & 15.67 & 78.1 & 69.2 & 0.932 & 3592.9 \\
% BO         & 62.18 & 53.80 & 54.89 & 69.3 & 1.110 & 4036.9  \\
% Ours w/o ICPI & 50.59 & 17.24 & 87.62 & 69.3 & 1.017 & 3288.3 \\
% Ours   & 51.39 & 18.69 & 108.6 & 69.4 & \textbf{1.104} & 4031.4 \\
% \bottomrule
% \end{tabular}
% }
% %\vspace{-0.4cm}
% \end{table*}

% \begin{table*}[tb]
% \centering
% \caption{Post-layout performance comparison on OTA2.}
% \label{tab:ota2}
% \scriptsize                  
% \setlength{\tabcolsep}{3pt}      
% \renewcommand{\arraystretch}{1.05}
% \resizebox{0.89\textwidth}{!}{%
% \begin{tabular}{lcccccc}
% \toprule
% Method & Gain (dB) & UGB (MHz) & CMRR (dB) & PM (deg) & FoM & Area ($\mu\text{m}^2$) \\
% \midrule
% Heuristic    & 53.15 & 1.454 & 55.70 & 59.4 & 0.894 & 7080.0 \\
% BO         & 62.14 & 1.610 & 62.80 & 70.4 & 1.022 & 8857.3 \\
% Ours w/o ICPI & 55.70 & 1.611 & 67.36 & 72.6 & 1.011 & 6490.5 \\
% Ours  & 62.72  & 1.597  & 80.86  & 72.3 & \textbf{1.098} & 6468.2 \\
% \bottomrule
% \end{tabular}
% }
% %\vspace{-0.4cm}
% \end{table*}

\begin{sloppypar}
In Table~\ref{tab:ota}, for each method we report the layout with the highest FoM found within the allowed simulation budget. Neither BO nor our ICPI framework directly optimizes area, and the objective is purely the electrical FoM, since the exposed parameter space in our current setup contains no explicit area controls. We therefore report area separately to distinguish gains obtained by enlarging the layout from gains achieved through more effective use of the exposed parameter space. Nonetheless, area or other multi-objective scores can be incorporated through the same feedback interface without modifying the loop.
\end{sloppypar}

\begin{sloppypar}
Furthermore, we observed a qualitative robustness difference. \textit{BO} can produce candidates that are geometrically legal and LVS-clean yet electrically non-functional once evaluated. On several unfavorable parameter combinations, excessive parasitics prevented the loop gain from reaching unity, and these committed candidates were assigned a zero FoM. In contrast, the infeasibilities encountered by \textit{Ours} arise as backend-infeasible intermediate proposals, such as over-constrained symmetry or overly aggressive wire widths that break legalization or routing. Such proposals are detected from failure logs and rolled back to a nearby feasible setting via the Executor and the design journal before commitment, so all committed layouts from \textit{Ours} and \textit{Ours w/o ICPI} are feasible, electrically functional, and LVS-clean. This contrast highlights the limitation of black-box methods that lack physical awareness under sparse simulation feedback.
\end{sloppypar}

\begin{sloppypar}
For OTA1, the parameter space is relatively modest, so BO can explore it reasonably well and reaches a FoM of $0.979$, slightly lower than \textit{Ours w/o ICPI} with $1.003$. Even in this smaller search space, however, Ours achieves the highest FoM $1.104$ with an area essentially identical to BO, showing that the act--observe--reflect ICPI loop can extract additional performance beyond both one-shot LLM guidance and black-box search. Although the ICPI layout is larger than the \textit{Heuristic} design, this is consistent with our setup, where area is reported separately rather than directly penalized in the objective. Notably, \textit{Ours w/o ICPI} already improves both performance and area over the \textit{Heuristic}, indicating that our framework can exploit the action space to enhance electrical behavior without necessarily increasing area.
\end{sloppypar}

\begin{sloppypar}
For OTA2, the action space is much larger, with a parameter dimension roughly two orders of magnitude higher than that of OTA1. Under the same 31-candidate optimization horizon, BO improves over the Heuristic on all four electrical metrics, but its CMRR remains below the target and its area increases by more than 25\%. Ours w/o ICPI improves both FoM and area over the Heuristic, although it does not meet the Gain and CMRR targets. In contrast, Ours meets all four electrical targets and achieves the highest FoM of $1.038$ while reducing area relative to both the Heuristic and BO layouts. Compared with Ours w/o ICPI, the full ICPI loop increases the FoM from $0.868$ to $1.038$ and slightly reduces area, while keeping UGB above its target.
\end{sloppypar}

\begin{figure}[tb]
  \centering
  \includegraphics[width=0.48\textwidth]{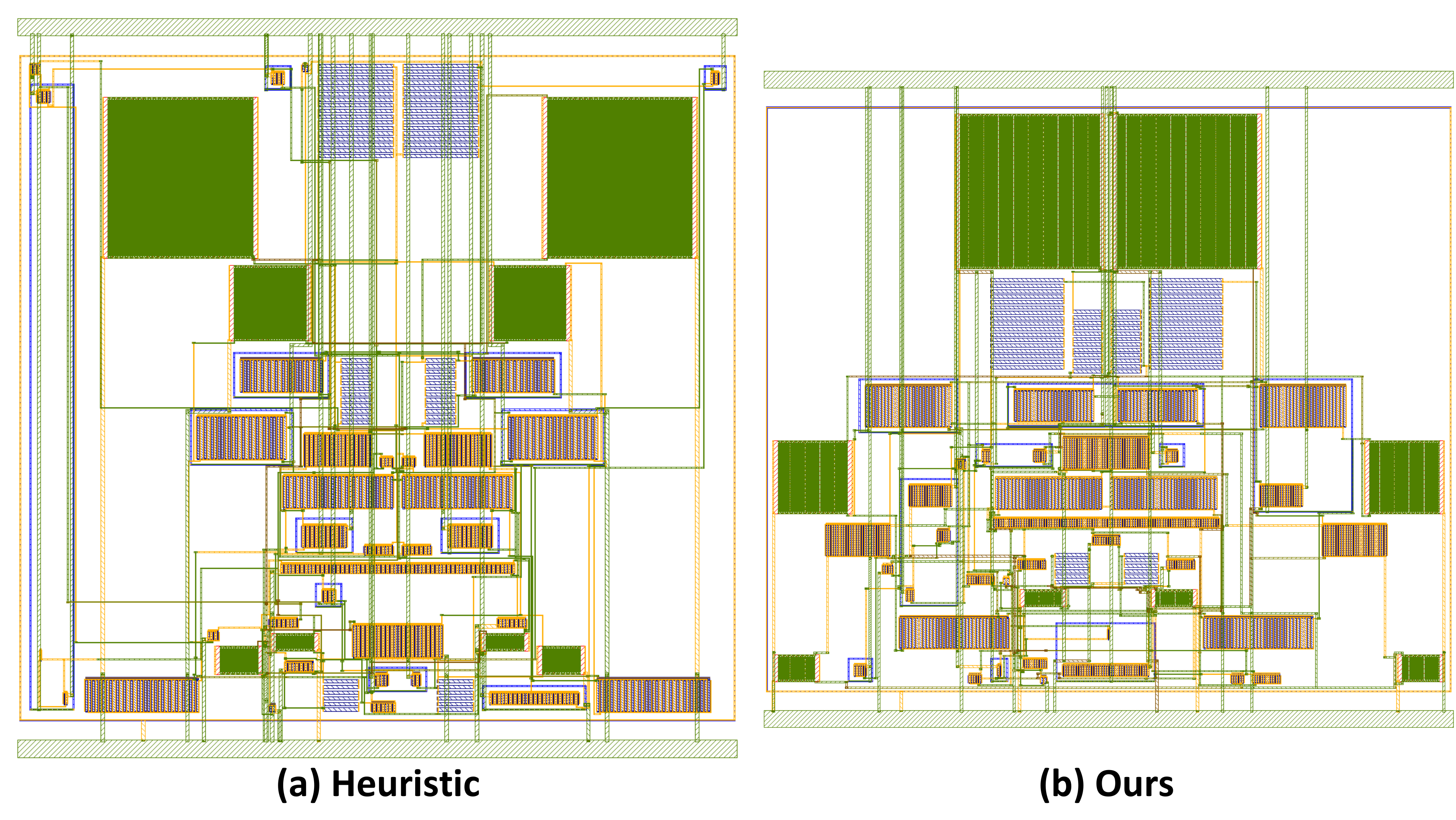}
  \caption{Layout comparison of (a) Heuristic and (b) Ours on OTA2.}
  \label{fig:layout}
  \vspace{-0.5cm}
\end{figure}

\subsection{Runtime and overhead}

To isolate the cost of the agent orchestration itself, we profile the agent loop with the same backbone LLM. Per round, orchestration takes about 37\,s and 12k tokens for OTA1 and about 43\,s and 17k tokens for OTA2, with input context accounting for over 90\% of the tokens, which amounts to roughly 18 to 21 minutes over 30 rounds. By comparison, parasitic extraction and the post-layout simulations alone require more than 141\,min per design under a conservative lower-bound timing, before accounting for place-and-route. The LLM orchestration therefore accounts for less than 16\% of total wall-clock time under this bound, and substantially less under realistic timing. The dominant cost remains the simulation-driven evaluation that the framework is designed to economize, rather than the agent loop.

\section{Conclusion}
\label{sec:Conclusion}
% ICLAD'26 submission
In this paper, we presented a simulation-aware LLM multi-agent framework that wraps a mature analog layout generator and performs in-context policy improvement over layout optimization parameters through an act--observe--reflect loop driven by sparse post-layout feedback. Experiments on real-world analog circuits show that, with only tens of post-layout simulations, our approach improves post-layout performance over the generator's built-in heuristics, BO-based tuning, and a non-ICPI baseline. While the framework is effective on the studied designs, scaling it to more complex systems remains challenging due to the growing complexity of state representation and the high cost of simulation-driven search.
Future work will address these limitations by exploring hierarchical partitioning and more scalable refinement strategies for larger circuits, and by extending the real-flow evaluation beyond the OTA benchmarks to a broader range of analog blocks.
To support reproducibility, we release a PDK-free demonstrator\footnote{\url{https://github.com/bingyang1132/ICLAD2026-demo-sim-aware-in-context}} that reproduces the core mechanisms of our framework in a synthetic analog-layout environment with a swappable LLM backbone.
Overall, this work establishes a viable state-aware refinement loop for simulation-driven analog layout optimization that can assist expert designers under tight evaluation budgets. 

\section{Acknowledgment}
%TODO
This work was supported in part by NSF under grant CCF-2112665, SRC under task 3160.007, Samsung, UT Austin’s iMAGiNE consortium, and an equipment donation from NVIDIA.

\clearpage
% \vspace{-.05in}
{
\scriptsize
% \footnotesize    
% \small
% \bibliographystyle{IEEEtran}
\bibliographystyle{acm}
\bibliography{./ref/TOP_sim.bib, ./ref/DAC24.bib}
}

\end{document}